\documentclass{article} 
\usepackage{iclr2023_conference_tinypaper,times}

\usepackage{amsmath,amsfonts,bm}

\def\eqref#1{equation~\ref{#1}}

\def\1{\bm{1}}

\DeclareMathAlphabet{\mathsfit}{\encodingdefault}{\sfdefault}{m}{sl}
\SetMathAlphabet{\mathsfit}{bold}{\encodingdefault}{\sfdefault}{bx}{n}

\usepackage{hyperref}
\usepackage{cleveref}
\usepackage{url}
\usepackage{caption}
\usepackage{subcaption}
\usepackage[pdftex]{graphicx}

\title{Native machine learning \\ for noisy microscopic data processing}

\author{Danil Afonchikov, Elena Kornaeva, Irina Makovik  \\
Orel State University \\
\texttt{danil.afonchikov@bk.ru, smkornaeva@gmail.com, irina.makovik@gmail.com}
\And
Alexey Kornaev \\
Innopolis University \\
\texttt{avkornaev@gmail.com}
}

\iclrfinalcopy 
\begin{document}

\maketitle

\begin{abstract}

Cells count become a challenging problem when the cells move in a continuous stream, and their boundaries are difficult for visual detection. To resolve this problem we modified the training and decision making processes using curriculum learning and multi-view predictions techniques, respectively. 

\end{abstract}

\section{Introduction}

There are two main approaches to low-quality images processing. The first approach deals with image restoration \citep{schermelleh2019super, lehtinen2018noise2noise, xu2020noisy}. This research is related to the second approach which deals with robustness of machine learning \citep{matiisen2019teacher, radosavovic2018data, Li2021Representation, bachman2019learning, wenzel2020hyperparameter}. Particulary, the research deals with {\it curriculum learning} \citep{wang2021survey} which is learning from simple tasks to complex ones, and {\it multi-view} post-processing which allows to make the final prediction based on multiple preliminary predictions.
The main application of the research is microscopic image and video processing. One of the ways to study blood cells inside a human body is capillaroscopy \citep{ingegnoli2018capillaroscopy}. Data on blood micro-circulations obtained using non-invasive methods of capillaroscopy normally have relatively low contrast since the observation takes place through soft tissues \citep{abdou2022deep,dremin2019dynamic}. Rare exceptions are the methods that require the use of the most sophisticated equipment and measurement procedures \citep{mckay2020imaging}. Some diseases are accompanied by a significant change in number of blood cells \citep{Deng2023}. An express {\it in vivo} blood count test that estimates at least low/high level of blood cells is required in hematology.         

\section{Methodology}
\label{sec:methodology}
The paper deals with an original synthetic video dataset of moving red and white blood cells (RBCs, WBCs, respectively). The proposed dataset is an attempt to connect the existing {\it{in vitro}} complete blood count tests (see \cref{sec:A_Data_collection}, \cref{fig:in_vitro_CBC}) ~\citep{BCCD_Dataset} and the prospective {\it{in vivo}} tests using capillaroscopy data (see \cref{sec:A_Data_collection}, ~\cref{fig:in_vivo_CBC}) ~\citep{herrick2021quantitative, gurov2018high}. Some examples of a nailfold capillaroscopy and synthetic dataset videos are given in the supplementary materials. The proposed synthetic dataset contains 1150 videos with 100 frames in each video of $3 \times 128\times 128$ frames. See \cref{sec:A_Data_collection} for detailed information on the dataset. 
The paper uses competence-based {\it curriculum learning} (CL) to reduce training time and increase accuracy of the models \citep{platanios2019competencebased}. The main idea of the method is to calculate difficulty of each sample and use the samples with difficulty score less or equal to the competence at a given training epoch (see \Cref{sec:CL_and_MV}, \Cref{eq:difficulty_competence}).
The proposed {\it multi-view} (MV) method refers to the idea of ensembling and using a set of slightly different networks that make a set of predictions to estimate the uncertainty of the predictions \citep{ashukha2020pitfalls}. But in work, a set of slightly different (augmented) variants of an input is used to make a set of preliminary predictions. It is supposed that the final prediction is the most frequent preliminary prediction, or prediction with the highest cumulative weight (see \Cref{sec:CL_and_MV}, \Cref{eq:MVMode}, \Cref{eq:cum_weights}, respectively).   

\section{Results and discussion}
\label{sec:results}

Two series of experiments were performed (see \Cref{sec:A_Data_collection}, \Cref{sec:details}).
The {\it first series} demonstrated that the proposed multi-view approach increases the accuracy significantly, which is partially the result of a relatively small crop area of the augmented frames (see \Cref{tab:simRes1Frame}). The number of multi-views and the parameters of data augmentation, e.g. random crop scale, are the additional hyperparameters of the multi-view approach. A set of preliminary tests demonstrated that the more the number of multi-views, the more the accuracy of the model. But, the dependence on the crop scale is not obvious and should be studied in further work. The best accuracy of $90.26\pm0.89\%$ was obtained with EffNet X3D, the network with relatively small number of trainable parameters, and the only network that processed a short 9 frame videos instead of single frames (see \Cref{tab:simRes1Frame}). ResNet18 had the highest number of trainable parameters, and the model demonstrated a tendency to overfitting which is the result of using a relatively small dataset (see \Cref{sec:details}, \Cref{fig:loss_and_accu_res}). The {\it the second series} dealt with a more difficult task, since the variance of RBCs distribution in the dataset is small compared to its mean value (see \Cref{sec:A_Data_collection}).Accuracy of $62\pm0.51\%$ with single-frame version of the dataset and accuracy of $65.6\pm0.52\%$ with 100-frames version of the dataset using 9 randomly selected consecutive frames were obtained. The effect of curriculum learning is not obvious, since the method increased the accuracy of the model for the video processing by about 1\%, but decreased the accuracy of the single-frame model. 
In both series, the use of videos of moving blood cells allowed to increase the accuracy of the models by about 2-4\%. The multi-veiw predictions demonstrated good results in all the tests. The method follows the principle of reductionism when the properties of a part correspond to the properties of the whole object. MV scans an image with relatively small random crops and accumulates the information on predictions, which is probably close to the idea of convolutional networks, but implemented at the post processing stage. The method was additionally tested in processing of noisy data in a public dataset (see \Cref{sec:additional_exp}).

\begin{table}[t]
\caption{Comparison of the baselines and the proposed models based on curriculum learning (CL) and multi-view (MV) predictions. The synthetic video dataset of moving blood cells was used.}
\label{tab:simRes1Frame}
\begin{center}
\begin{tabular}{lccccc}
\multicolumn{1}{c}{\bf MODEL}  & {\bf \#PARAM.}   & {\bf \#EP.} & {\bf INPUT} & \multicolumn{2}{c}{\bf ACCURACY, \%} \\
                               & {(train. par.)}  &             &             & Baseline  & MV / CL         \\
\hline \\
\multicolumn{6}{c}{\it White blood cells (WBC)} \\
EffNet pr.              & 3.8 M (all)        & 400 & [9,3,64,64] & $\bf 83.56\pm3.05$ & $\bf 90.26\pm0.89 / -$   \\
9-layer CNN                    & 4.4 M (all)        & 800 & [3,32,32]   & $81.57\pm3.32$ & {$89.39\pm2.58 / -$} \\
ResNet18 pr.                & 11.2 M (all)       & 400 & [3,256,256] & $82.61\pm2.53$ & $87.83\pm2.70 / -$  \\
ViT pr.                     & 303 M (2 K)        & 400 & [3,224,224] & $76.31\pm2.76$ & $85.00\pm2.15 / -$       \\
\multicolumn{6}{c}{\it Red blood cells (RBC)} \\
EffNet pr.              & 3.8 M (all)        & 400 &[3,64,64]& $62.6\pm0.53$ & $- / 62\pm0.51$   \\
EffNet pr.              & 3.8 M (all)        & 200 &[9,3,64,64]& $64.6\pm0.52$ & $- / 65.6\pm0.52$   \\

\end{tabular}
\end{center}
\end{table}

\section{Conclusion}
Some of the real-world cells datasets are not applicable for cells detection and labeling, but the task of qualitative or quantitative estimation of the cells number remains actual. The obtained results demonstrated that the number of relatively big and low concentrated white blood cells can be estimated as high or low with the accuracy of about 90\%. But the number of relatively small and high concentrated red blood cells cannot be estimated accurately, the obtained result in accuracy of about 66\% is close to random predictions. The methods of curriculum learning and multi-view predictions are easy in implementation and promising in application to processing of relatively small and low quality datasets.  
The further research should be related to processing of the real-world data obtained with {\it in vivo} methods of blood cells visualization. Data collection for a case of well-separated blood count test results and solution of the binary classification problems are of interests.   

\subsection*{Acknowledgement}
This research has been financially supported by The Analytical Center for the Government of the Russian Federation (Agreement No. 70-2021-00143 dd. 01.11.2021, IGK 000000D730324P540002). Authors also express gratitude to R\&D Center of Biomedical Photonics at Orel State University for sharing their knowledge in the field of microscopic imaging.

\subsubsection*{URM Statement}
We acknowledge that all the authors meet the URM criteria of ICLR 2024 Tiny Papers Track.

\bibliography{iclr2023_conference_tinypaper}
\bibliographystyle{iclr2023_conference_tinypaper}

\appendix
\section{Data collection}
\label{sec:A_Data_collection}

Modern complete blood count (CBC) datasets are based on {\it in vitro} tests results. The datasets contain high quality images of blood cells. So, object detection methods can be applied to process them. Data obtained with {\it in vivo} videocapillaroscopy contain images distorted with blur and noise. So object detection methods are not normally applicable. In addition, such distortions worsen the results of AI model predictions, making said models inadequate. 

It is supposed that implementation of native machine learning methods including curriculum learning and multi-view predictions in practice of {\it in vivo} CBC tests can make predictions more reliable. In order to test that, the dataset of blood cells with increasing difficulty factor is needed. A synthetic video dataset of moving blood cells was generated to approach the solution of a problem of {\it in vivo} CBC tests.

The dataset includes clear, blurred, noisy, and both blurred and noisy samples in proportion $0.130 \times 0.304 \times 0.217 \times 0.349$, respectively. The number RBCs and WBCs have normal distribution in the dataset with mean values of 5000 and 202 and standard deviations of 97.9 and 95.5 respectively. 'High' RBCs or 'high' WBCs levels are used as class '1' in labels. It is supposed that the 'high' level corresponds to the number of cells higher than the mean value. The dataset contains 576 (0.5) videos with 'high' WBCs and 593 (0.512) videos with 'high' RBC levels. 

To simulate Brownian motion of blood cells, the Wiener process equation is applied:
\begin{equation}
    \label{eq:wiener}
    W\bigg(\frac{i}{n} \bigg) = W\bigg(\frac{i-1}{n}\bigg) + \frac{Y_i}{\sqrt{n}}.  
\end{equation}
where $n$ is the number of video frames, $i$ is the index of current frame, $Y_i = \pm 1$ , determined randomly.

\begin{figure}[t]
     \centering
     \begin{subfigure}[b]{0.19\textwidth}
         \centering
         \includegraphics[width=\textwidth]{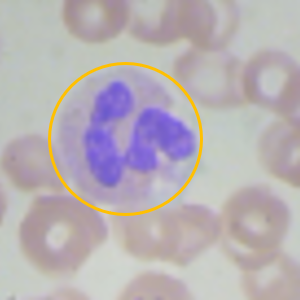}
         \caption{in vitro}
         \label{fig:in_vitro_CBC}
     \end{subfigure}
     \hfill
     \centering
     \begin{subfigure}[b]{0.19\textwidth}
         \centering
         \includegraphics[width=\textwidth]{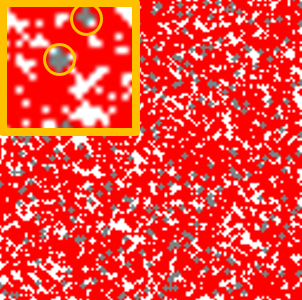}
         \caption{synthetic, $b = 0$}
         \label{fig:synth_b0}
     \end{subfigure}
     \hfill
     \centering
     \begin{subfigure}[b]{0.19\textwidth}
         \centering
         \includegraphics[width=\textwidth]{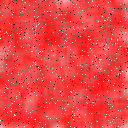}
         \caption{synthetic, $b = 5$}
         \label{fig:synth_b5}
     \end{subfigure}
     \hfill          
     \begin{subfigure}[b]{0.19\textwidth}
         \centering
         \includegraphics[width=\textwidth]{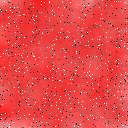}
         \caption{synthetic, $b = 10$}
         \label{fig:synth_b10}
     \end{subfigure}
     \hfill
     \centering
     \begin{subfigure}[b]{0.19\textwidth}
         \centering
         \includegraphics[width=\textwidth]{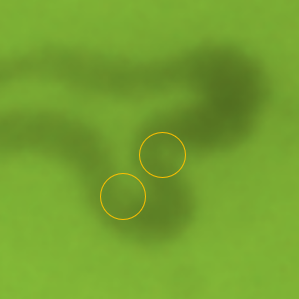}
         \caption{in vivo}
         \label{fig:in_vivo_CBC}
     \end{subfigure}
     \hfill
    \caption{Blood cells microscopy: moving along the path from {\it{in vitro}} (a) to {\it{in vivo}} (e) blood count tests using the synthetic blood count dataset with different values of blurring $b$ in the samples (b)-(d). Some of the white blood cells situated among the red blood cells are marked with circles (b), (e).}
    \label{fig:CBCdatasets}
\end{figure}

\Cref{eq:wiener} was applied to each coordinate and each cell to determine the moving cell's position at each video frame. Then the box blur was applied and the noise was drawn (see \Cref{fig:CBCdatasets}).  

\section{Curriculum learning and multi-view predictions intuition}
\label{sec:CL_and_MV}

{\it Curriculum learning}

Curriculum learning is a machine learning strategy when the model is trained on a sequence of samples and the difficulty of each sample gradually increases. The idea behind this approach is to mimic the way humans learn, by starting with simple concepts and gradually building up to more complex ones. This can help the model to learn more effectively and efficiently, as it is able to build upon its existing knowledge as it progresses through the curriculum.

Competence-based curriculum learning is a specific type of curriculum learning. The competence of the model is evaluated, and in the current learning epoch the samples with difficulty ($d$) less or equal to competence score ($c(t)$) are used \citep{platanios2019competencebased}. The difficulty estimation ($d$) and the competence equation ($c(t)$) have the following formalization:
\begin{equation}
    \label{eq:difficulty_competence}
    d = \alpha b + \beta l, \quad c(t) = \min \bigg( 1, \sqrt[^p]{t \frac{1 - c_0^p}{T} + c_0^p} \bigg), 
\end{equation}
where $b$ is the blur radius for box blurring, $(0 \leq b \leq 10)$, $l$ is the distance from the given target value to the mean value (exact number of cells), $\alpha$ and $\beta$ are the hyperparameters $(0 \leq \alpha, \beta \leq 1, \alpha + \beta = 1)$, $c_0$ is the initial competence value, $T$ is the total duration of curriculum learning phase, $t$ is current epoch, $p \geq 1$.

\begin{figure}[h]
    \begin{center}
    \includegraphics[width=0.8\textwidth]{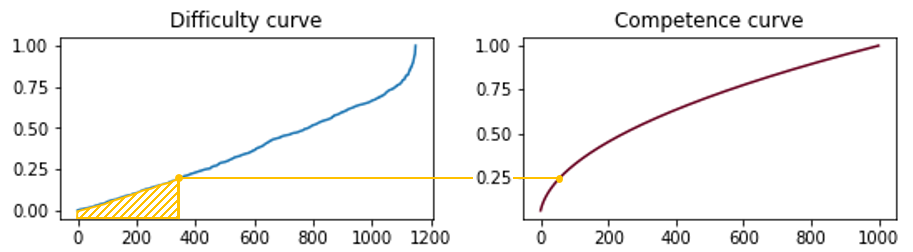}
    \end{center}
    \caption{Curriculum learning intuition. The difficulty and competence curves represents \Cref{eq:difficulty_competence} when $\alpha = \beta = 0.5$, $c_0 = 0.05$, $T = 1000$, and $p = 2$.}
\end{figure}

\begin{figure}[t]
    \centering
    \includegraphics[width=0.98\textwidth]{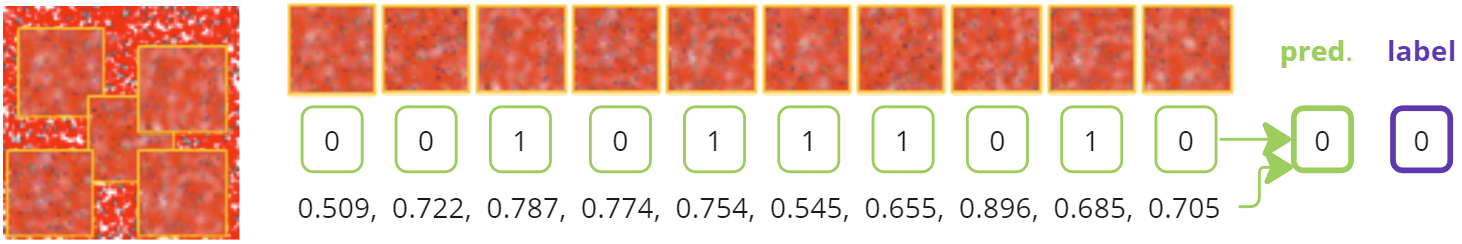}
    \caption{Multi-view intuition. The network recieves a set of fragments of the sample and makes a set of predictions. The final prediction corresponds to the mode value (MVM, see \Cref{eq:MVMode}), or to the bin count weighted with confidences (MVWCo-S, see \Cref{eq:cum_weights}).}
    \label{fig:multi_view}
\end{figure}

{\it Multi-view}

Microscopic images of cells normally include a number of cells and a small part of an image contains approximately the same information on cells condition and concentration as the whole image. So, data augmentation (image cropping, first of all) expands the dataset during the training and makes a small dataset enough for deep learning models. But also the augmentation allows to average test predictions and estimate their uncertainty using techniques of multi-view predictions. 

The goal of multi-view predictions is to use a sample with augmentations multiple times as follows. The inference model receives $m$ augmented copies of each input and calculates $m$ pairs of predicted classes and {\it confidences} of the predictions $\{ y^{(i,j)}\:', c^{(i,j)} \}$ for each input, respectively:  

\begin{equation}
    \label{eq:con}
        y^{(i,j)}\:' =  \underset{k}{\arg\max} \left[ h_k^{(i,j)} \right], \quad c^{(i,j)} = \underset{k} max({h}_k^{(i,j)}),
\end{equation}
where $k$ is the class number, $(i,j)$ are the dummy indexes that refer to $j$\textsuperscript{th} augmented copy of $i$\textsuperscript{th} sample, $c^{(i,j)}$ is the maximum component of a model prediction vector $\mathbf{h}^{(i,j)}$ which is called confidence,  $0 < c^{(i,j)} < 1$. 

The confidence values $c^{(i,j)}$ can be interpreted as an certainty or uncertainty estimator of the model predictions \citep{Pearce2021}: the more the confidence value, the more the certainty of the prediction.

There are a few ways to calculate resulting predictions. The simplest way is to calculate mode values over the multi-view predictions (MVM): 

\begin{equation}
    \label{eq:MVMode}
        y^{(i)}\:' =  \underset{j}{mode} \left[ y^{(i,j)}\:' \right].
\end{equation}

The alternative approach uses information on certainty of the predictions and deals with cumulative confidences:

\begin{equation}
    \label{eq:cum_weights}
    z_{k = y^{(i,j)}\:'}^{(i)} =  \sum_{j}{c^{(i,j)}}, \quad y^{(i)}\:' = \underset{k}{\max}{(z_k^{(i)})},
\end{equation}
where $\mathbf{z}^{(i)}$ is the additional vector with components $z_k^{(i)}$ which accumulates weights of the predicted classed over the multi-views.  

\Cref{eq:cum_weights} represents bin count of predicted classes ${y^{(i,j)}\:'}$ with weights ${c^{(i,j)}}$. The method is called soft multi-view predictions weighted with confidences $c^{(i,j)}$ (MVWCo-S).   

\section{Details of experiments}
\label{sec:details}

Two series of experiments were performed. In both series the dataset was divided into training, validation and test sets in the proportions of 0.6, 0.2, 0.2, respectively. In both series the random resized crop of $10-20\%$ of the frame area was applied for the input data augmentation at the training, validation, and test stages. The Adam optimizer (momentum 0.9) was used with an initial learning rate of $0.001$. Five parallel tests with random seeds of $[42, 0, 17, 9, 3]$ were performed, the means and standard deviations were calculated. 

{\it The first series of the experiments} dealt with various network architectures and with classification of 'high' or 'low' levels of WBCs (higher or lower than the mean value of WBCs). Four network architectures were applied with the following training tactics (see \cref{tab:simRes1Frame}):
\begin{itemize}
    \item pretrained EffNet X3D with $3.8$ million parameters (fine tuned);
    \item 9-layer CNN \citep{Xia2021} with $4.4$ million parameters (randomly initialized);
    \item pretrained ResNet18 with $11.2$ million parameters (fine tuned);
    \item pretrained ViT L 16 with $303$ million parameters (2 thousand parameters of the 'head' of the model were randomly initialized, the rest of the parameters were frozen).
 \end{itemize}

\begin{figure}[t]
     \centering
     \begin{subfigure}[b]{0.24\textwidth}
         \centering
         \includegraphics[width=\textwidth]{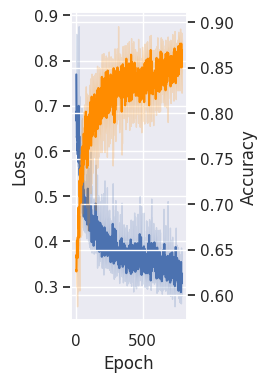}
         \caption{9-layer CNN}
         \label{fig:in_vitro_CBC}
     \end{subfigure}
     \hfill
     \centering
     \begin{subfigure}[b]{0.24\textwidth}
         \centering
         \includegraphics[width=\textwidth]{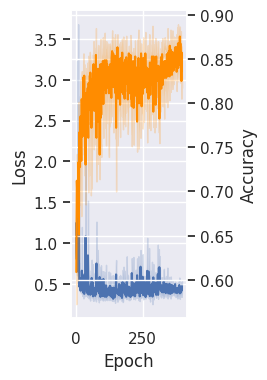}
         \caption{ResNet18 pretr.}
         \label{fig:synth_b0}
     \end{subfigure}
     \hfill
     \centering
     \begin{subfigure}[b]{0.24\textwidth}
         \centering
         \includegraphics[width=\textwidth]{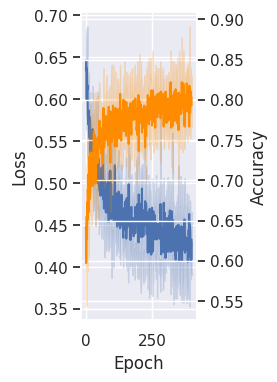}
         \caption{ViT pretr.}
         \label{fig:synth_b5}
     \end{subfigure}
     \hfill          
     \begin{subfigure}[b]{0.24\textwidth}
         \centering
         \includegraphics[width=\textwidth]{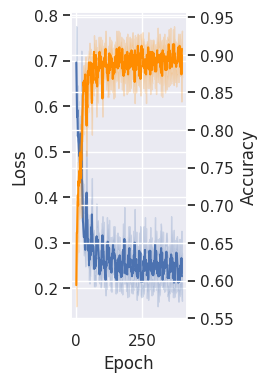}
         \caption{EffNet X3D pretr.}
         \label{fig:synth_b10}
     \end{subfigure}
     \hfill
    \caption{The validation loss (blue lines) and the validation accuracy (orange lines) functions (see \Cref{tab:simRes1Frame}) while the train loss functions demonstrated monotonous decreasing in all the experiments.}
    \label{fig:loss_and_accu_res}
\end{figure}

Ten multi-views (augmented versions) of a test sample were used (see \cref{fig:multi_view}). \Cref{tab:simRes1Frame} and \Cref{fig:loss_and_accu_res} demonstrate the experiments results and the training processes, respectively. 

{\it The second series of the experiments} dealt with both single-frame version and 100-frames version of the dataset with the classification of 'high' or 'low' level of RBCs. Pretrained Effitient X3D model \citep{feichtenhofer2020x3d} was used. The model was trained with cosine annealing learning rate with starting value of $0.001$. Competence-based curriculum learning \citep{platanios2019competencebased} was used. The model was trained in 400 epochs with single-frame version of the dataset and in 200 epochs with 100-frames version of the dataset using 9 randomly selected consecutive frames (see \Cref{tab:simRes1Frame}).

\section{Additional experiments on CIFAR-10 dataset}
\label{sec:additional_exp}
Since the proposed multi-view approach demonstrated stable and positive influence on the accuracy of the models (see \Cref{tab:simRes1Frame}), the approach was additionally tested with CIFAR-10 dataset \citep{Krizhevsky2009LearningML}. The dataset was manually corrupted with {\it asymmetric noise} in labels \citep{Xia2021, ma2020normalized}. Both, in the main experiments (see \Cref{tab:simRes1Frame}) and in this experiments the 9-layer CNN network presented in \citep{han2018coteaching, Xia2021} was used. For all experiments, the Adam optimizer with momentum of 0.9 and weight decay of $0.01$ was used, the batch size was set to 128, the models were trained in 200 epochs with an initial rate of $0.001$ and a schedule in correspondence to paper by \citet{Xia2021}. Unlike the paper by \citet{Xia2021} the inference models correspond to the best validation loss. The models were initialized with the random weights before the training. Each experiment was performed within 5 parallel tests with random seeds of $[42, 0, 17, 9, 3]$. Unlike the main experiments of this paper, the technique of label smoothing was applied with noisy labels \citep{wei2021smooth} in this series of experiments. The combination of label smoothing and multi-view predictions allowed the model to exceed some of the results of other authors (see \Cref{tab:simResCIFAR10}). The number of multi-views was set to 50.

\begin{table}[t]
\caption{Comparison of the proposed multi-view (MV) approach with alternative approaches to noisy data processing. CIFAR-10 dataset and its corrupted version with noise in labels were used. Top two results are highlighted.}
\label{tab:simResCIFAR10}
\begin{center}
\begin{tabular}{lccccc}
\multicolumn{1}{c}{\bf MODEL}  & {\bf LS }     & {\bf \#PAR.}     &  {\bf \#EP.} & \multicolumn{2}{c}{\bf ACCURACY, \%} \\
                               &   ratio       & {(train. par.)}  &              & Clean labels  & 20\% asym. noise \\
\hline \\
\multicolumn{6}{c}{\it Proposed models} \\
9-layer CNN              & 0 / 0.4 & 4.4 M (all) & 200 & $87.19\pm1.03$ & $80.67\pm0.52$ \\
9-layer CNN + MVM        & 0 / 0.4 & 4.4 M (all) & 200 & $\bf 92.28\pm0.65$ & $87.55\pm0.22$ \\
9-layer CNN + MVWCo-S    & 0 / 0.4 & 4.4 M (all) & 200 & $\bf 92.34\pm0.66$ & $\bf 87.60\pm0.23$ \\
\multicolumn{6}{c}{\it \citet{Xia2021}} \\
9-layer CNN + Co-teach.  & - & 4.4 M (all) & 200 & - & $83.87\pm0.24$ \\
9-layer CNN + JoCor      & - & 4.4 M (all) & 200 & - & $80.96\pm0.25$ \\
9-layer CNN + CNLCU-S    & - & 4.4 M (all) & 200 & - & $85.06\pm0.17$ \\
\multicolumn{6}{c}{\it \citet{wei2021smooth}} \\
ResNet34 + LS            & 0 / 0.4 & - & 200 & $91.44\pm0.16$ & $\bf 90.49\pm0.1$ \\
\end{tabular}
\end{center}
\end{table}

\end{document}